\newif\ifcomments  
\commentsfalse

\documentclass[a4paper,table]{article}
\usepackage{INTERSPEECH2023}
\usepackage[dvipsnames]{xcolor}
\usepackage{highlight2}
\usepackage{comment}
\usepackage{hyperref}
\usepackage{url}
\usepackage{graphicx}
\usepackage{enumitem}

\newcommand{\wrh}[1]{\ifcomments\textcolor{red}{\textbf{wrh: #1}}\else \ignorespaces \fi}

\newcommand{\eos}{\texttt{<EOS>}}

\title{Semantic Segmentation with Bidirectional Language Models Improves Long-form ASR}
\name{W. Ronny Huang$^*$, Hao Zhang$^*$, 
Shankar Kumar\sthanks{\;\;Equal contribution} , Shuo-yiin Chang, Tara N. Sainath}

\address{Google Research, USA}
\email{\{wrh, haozhang, shankarkumar, shuoyiin, tsainath\}@google.com}

\begin{document}
\maketitle
\begin{abstract}
We propose a method of segmenting long-form speech by separating semantically complete sentences within the utterance.
This prevents the ASR decoder from needlessly processing faraway context
while also preventing it from missing relevant context within the current sentence.
Semantically complete sentence boundaries are typically demarcated by punctuation in written text;
but unfortunately, spoken real-world utterances rarely contain punctuation.
We address this limitation by distilling punctuation knowledge from a bidirectional teacher language model (LM)
trained on written, punctuated text.
We compare our segmenter, which is distilled from the LM teacher, against a segmenter distilled from a acoustic-pause-based teacher used in other works, on a streaming ASR pipeline.
The pipeline with our segmenter achieves a 3.2\% relative WER gain along with a 60 ms median end-of-segment latency reduction on a YouTube captioning task.

\end{abstract}

\section{Introduction}
\label{sec:intro}

Recognizing long-form speech (tens of minutes) in short segments of a few seconds is a common practice
for improving ASR accuracy and user-perceived latency \cite{paturi2021directed,shangguan2021dissecting}.
Model state is wholly or partially discarded at segment boundaries,
helping to prevent the model from entering strange states unseen during training (which is typically short-form)
and making room for more diversity in the beam search hypotheses \cite{prabhavalkar2021less}.
Conventional end-of-segment (EOS) classifiers have relied primarily on acoustic signals, such as long silences,
detected directly during inference by a voice activity detector (VAD).
Long silences, however, may not always demarcate semantically complete sentences,
as speakers often have mid-sentence hesitations in real world speech.

Recent work on end-to-end (E2E) segmentation allows the E2E model itself to predict EOS boundaries
during decoding in an online, frame-synchronous manner \cite{huang2022e2e,huang2022e2esegmentation}.
The benefit is that the E2E model conditions on both the audio and decoded text,
and thus can fuse this information together to make better EOS decisions.
However, the segmentation quality depends not only on the context available to the model,
but also on the training signal for where EOS \textit{should} occur. 
Previous work relied on a pause detector teacher which decrees EOS placement between two words
whenever there is a long silence between them \cite{huang2022e2e,huang2022e2esegmentation}.
The exact threshold of the silence length depends on a set of rules regarding the surrounding words \cite{chang2022turn},
but the EOS training signal is still primarily driven by acoustic silence.

Here, we opt instead for an EOS training signal that is based explicitly on semantically relevant boundaries.
In written text, semantically complete sentences are explicitly delimited by punctuation,
such as commas, periods, or question/exclamation marks.
Such punctuation is a natural fit for the EOS training signal,
but unfortunately spoken utterances typically do not contain punctuation.
For example, instead of saying ``hi ivy period come here period'',
users simply say ``hi ivy come here'',
even though the periods would be present in written form.
It's possible to hand-transcribe punctuation marks,
but third-party punctuation annotations are typically less reliable and more costly.

In this work, we contribute a novel solution to this problem of obtaining semantic boundary signals for training an ASR segmenter.
Our approach is to distill the knowledge of punctuation from the written domain to the spoken domain.
We do this by first training a bidirectional language model (LM) on a separate corpus of written domain text and using it as a teacher model.
The LM is specifically tasked to predict the punctuation marks from the corpus.
Second, we run the LM teacher on all of our ASR training data to impute punctuation on all the transcripts.
Finally, we train the segmenter on the punctuation-injected ASR training data.
Note that this is the same procedure used in prior work with the pause detector teacher,
except that in the second step our LM teacher replaces their pause detector.
With this approach, we achieve a 3.2\% WER gain relative to the pause detector teacher-distilled model,
along with a 60 ms median EOS latency reduction,
on a production single-pass RNN-T ASR model.


\subsection{Related work}
Many prior works have applied semantic segmentation and punctuation prediction in online settings
\cite{vandeghinste2023fullstop,hlubik2020inserting,vsvec2021transformer}.
\cite{behre2022smart} mixes acoustic and streaming LM features, along with a one-word look-ahead,
to improve segmentation quality in streaming settings,
while \cite{behre2023streaming} uses a LSTM-based punctuation tagging model, similar to our LM teacher,
applied directly during inference with dynamic decoding windows.
Both of these approaches present potential challenges in latency critical conditions,
while our work reduces latency by design.
\cite{zhou2022punctuation} applies a similar joint multitask training framework for predicting the punctuation assuming that ground truth punctuation is available.
Our work relaxes this assumption which makes it more applicable to general scenarios.
Finally, \cite{mccarthy2022} used a very similar LM-based segmenter distillation technique for machine translation;
we explore applying such a technique to streaming ASR.

\section{Method}
\label{sec:method}

\subsection{Joint segmenting and decoding with E2E model}
\begin{figure}[]
  \centering
  \includegraphics[width=0.99\linewidth]{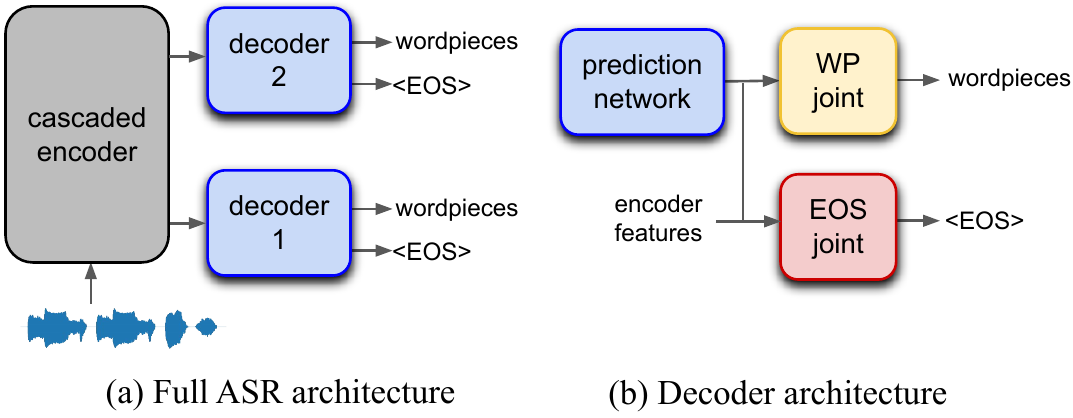}
  \caption{Cascaded encoder ASR architecture. WP: wordpiece. EOS: end-of-segment.}
  \label{fig:architecture}
\end{figure}

Our base E2E model, architecture in Figure \ref{fig:architecture},
employs a streaming RNN-T architecture whose joint network (yellow) predicts wordpiece posteriors and conditions on prediction network (blue) and encoder features.
Instead of having the segmenter be a separate model,
we allow the E2E model to jointly predict wordpieces \textit{and} segment boundaries by following the design in \cite{huang2022e2e}.
In particular, we augment the RNN-T architecture with an auxiliary head (red), called the EOS joint network,
whose shape, inputs, and outputs matches that of the original, wordpiece (WP) joint network,
as shown in Figure \ref{fig:architecture}(b).
This network's role is to predict the posterior probability of the end-of-segment, or \eos{},
which signals placing a segment boundary at the current frame.
In order to prevent degrading WER, we fine-tune the EOS joint network while holding the weights of the base RNN-T frozen.
During fine-tuning, the training data transcripts contain \eos{} tokens between semantically complete sentences,
whereas training of the base network does not.

Since semantically complete sentences are typically demarcated by punctuation such as commas, periods, or question/exclamation marks,
they are the natural supervision signal on which to place \eos{} tokens.
Unfortunately spoken domain utterances typically do not contain punctuation,
or the punctuation is provided by a transcriber rather than original speaker, which can be unreliable.
We address this problem by training a bidirectional language model on written domain corpora containing punctuation,
and applying that language model to our spoken domain transcripts to insert \eos{} tokens.
We henceforth call the language model the ``LM teacher'' and the EOS joint the ``student''.
Next, we discuss the details of the language model.

\begin{figure}[]
  \centering
  \includegraphics[width=0.9\linewidth]{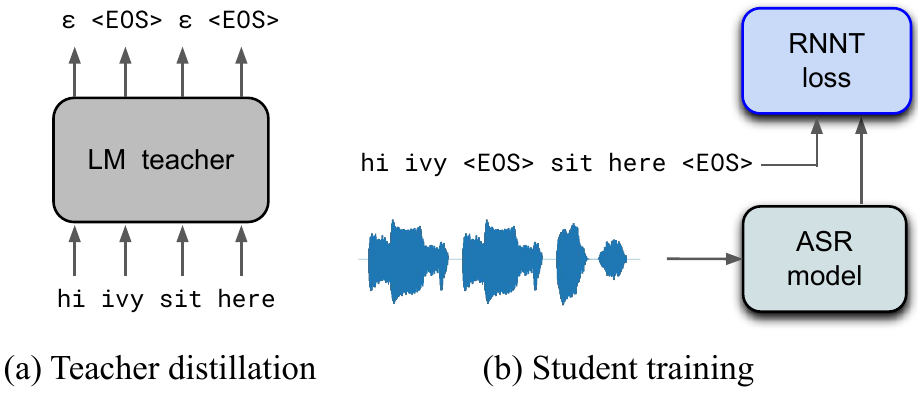}
  \caption{Annotating training data with \eos{} obtained via distillation from LM teacher.}
  \label{fig:distillation}
\vspace{-10pt}
\end{figure}

\subsection{Language model teacher}
We use an autoregressive recurrent neural network model, architecture identical to \cite{mccarthy2022}, for semantic (complete sentence) segmentation. The model has an encoder-decoder architecture. The encoder is bidirectional to fully utilize the context on both sides of a token under consideration. The decoder is autoregressive so as to condition predictions on the entire history of complete sentences. Given an input text window containing $w$ tokens: $\mathbf{x}=x_1,\ldots,x_{w}$, the model predicts an output sequence $\mathbf{y}$ of the same length: $\mathbf{y}=y_1,\ldots,y_{w}$, $y_i$ $\in$ \{$\epsilon$,\eos{}\} with probability,
\[
\arraycolsep=1.0pt\def\arraystretch{1.5}
\begin{array}{rcl}
  p_\theta(\mathbf{y} \mid \mathbf{x}) &=& \prod_{t=1}^w p_\theta(y_{t} \mid \mathbf{y}_{<t}, \mathbf{x})  \\
                                      &:=& \prod_{t=1}^w p_\theta(y_{t} \mid \mathbf{y}_{<t}, \mathbf{BiRNN(x)}_t). \\
\end{array}
\]
Figure~\ref{fig:distillation}(a) shows an example input output pair.

\subsection{Distillation and student fine-tuning}
Once the \eos{} positions are identified by the teacher, they are injected into the training transcripts between the appropriate words, as shown in Figure \ref{fig:distillation}(b).
The EOS joint network student is then fine-tuned via the RNN-T loss to predict all the label tokens, both \eos{} and wordpieces alike.
It is important to have the EOS joint predict the wordpieces though their posteriors are unused.
This can be seen as a multi-task objective to ensure that the \eos{} tokens are emitted
only after the appropriate wordpieces are emitted.
We also train with FastEmit regularization \cite{yu2021fastemit} to encourage the \eos{} token to be emitted as quickly as possible after the end of the last word in each segment.

\subsection{Inference}
During inference, a standard beam search is run using the wordpiece posteriors
emitted by the wordpiece joint network.
Meanwhile, the EOS joint network emits an \eos{} posterior at each frame.
There is no dependency between the tokens emitted by the wordpiece joint and EOS joint.
When the negative log \eos{} posterior falls below a preset threshold---the EOS threshold---a segment boundary is emitted.
Typically, all states are reset at segment boundaries,
but here we pass the encoder state to the next segment as well as the decoder state of the top hypothesis.
The beam search is reset, however, and all the hypotheses are discarded.
The top hypothesis from the prior segment is \textit{finalized} and passed to downstream tasks.
Often times, the finalization latency, or EOS latency, is an important metric for user experience,
because it directly influences how long it takes to execute the downstream task after the user has spoken.
We measure the EOS latency in our experiments in Table \ref{tab:main}.
\section{Setup}
\label{sec:setup}

\subsection{Datasets}
\label{sec:datasets}
We are aware of the sensitive nature of ASR research and other AI technologies used in this work.
The training set for the ASR model consists of 400M anonymized hand- or machine-transcribed utterances from various domains,
such as voice search, farfield, dictation, telephony, and YouTube.
All the YouTube utterances are machine transcribed.
The data is diversified via multi-style training \cite{kim2017mtr}, random down-sampling from 16 to 8 kHz \cite{li2012improving}, and SpecAug \cite{Park2019}.


The training set used for the LM teacher is the English C4 data set \cite{raffel2020}. On the C4 paragraphs, we apply a fixed set of inference rules to identify sentence-terminal punctuation marks. These rules disambiguate punctuation marks that can be used internally as well as externally in sentences, such as ``.'' in ``XYZ Inc. is a public company.''. The sentence-terminal punctuation marks are converted to \eos{}. The formatting information that is not present in ASR model output such as casing and sentence-internal punctuation is removed.
Finally, the paragraphs are chunked into overlapping text snippets of at most 40 words each with an overlapping window of size 10. This sliding window data preprocessing step improves training efficiency as well as model robustness.

The evaluation is performed on a YouTube test set, YT\_LONG,
which consists of sampled videos from YouTube video-on-demand.
YT\_LONG has 77 total utterances consisting of 22.2 hours and 207,191 words, with a median utterance length of 14.8 minutes.
YouTube videos are diverse in terms of content, which makes YouTube captioning an ideal task for evaluation.

\subsection{Architecture}
The ASR architecture used in this work, shown in Figure \ref{fig:architecture}(a),
is a cascaded encoder model with two separate decoders.
It is similar to that of \cite{ding2022unified},
with the same vocabulary, architecture, optimizer setup, and training data.
The model consists of a causal encoder, cascaded encoder, with parameters of 47M, 60M, respectively. 
The causal encoder consists of only causal convolutions and left-context self attention.
The cascaded encoder inputs the feature vector sequence from the causal encoder,
and it consists of convolutions and self-attention with a total right-context of 900 ms across all its layers.
Separate, identical decoders consume the outputs of each encoder.
Each decoder (Figure \ref{fig:architecture}(b)) consists a joint network which predicts wordpieces and one which predicts \eos{} tokens,
totaling about 4.4M parameters.
Both joint networks condition on the concatenation of encoder features and prediction network features.
The prediction network is a stateless prediction network \cite{botros2021tied} with a context window of 2 tokens.
The optimizer used is Adam with a transformer learning rate schedule and exponential-moving-average-stabilized gradient updates.
The model, implemented in Lingvo \cite{shen2019lingvo}, is initially trained on 64 TPU chips with a global batch size of 4096 for 500k steps.
We fine-tune the EOS joint network for 50k steps on only the YouTube portion of the training data (Section \ref{sec:datasets}).

\subsection{Three operating modes}
\label{sec:modes}
Figure \ref{fig:architecture}(a) shows that there are two separate decoders,
each independently emitting its own wordpieces and EOS tokens.
Given this fact, there are three modes by which the model can operate.
They are summarized in Table \ref{tab:modes} and described as follows.

\begin{enumerate}[leftmargin=*]
    \item \textbf{1st pass}: Segment boundaries and wordpieces are determined by causal Decoder 1.
    This mode is effectively identical to a single-pass causal E2E model,
    which is the most common model in the literature and also the most straightforward.
    It has the advantage of quick EOS and wordpiece decisions,
    but has lower quality due to lack of right context.
    \item \textbf{EOS2-segmented 2nd pass}: Segment boundaries and wordpieces are determined by non-causal Decoder 2.
    This mode is effectively identical to a single-pass non-causal E2E model,
    It is slower to emit EOS and wordpieces, but has higher quality due to the extra right context.
    This mode is used for non-latency-critical applications,
    such as offline batch processing.
    \item \textbf{EOS1-segmented 2nd pass}: Segment boundaries determined by Decoder 1
    are used to segment the beam search output of Decoder 2, from which the wordpieces are determined.
    This mode runs both passes simultaneously.
    When Decoder 1's EOS is emitted at frame $t$, Decoder 2's beam search is reset.
    While more complicated and unconventional of an approach,
    the purpose of this method is to achieve fast EOS (the EOS latency is identical to that of the 1st pass mode) and high quality simultaneously.
    To achieve the low EOS latency,
    we must apply a dummy right-context frame injection technique 
    to prevent Decoder 2's beam search from waiting for 900 ms of context beyond frame $t$.
    Details of this technique are discussed in \cite{huang2022e2esegmentation}.
\end{enumerate}

\begin{table}[]
\caption{Operating modes for the ASR model}
\centering
\label{tab:modes}
\resizebox{.8\columnwidth}{!}{%
\begin{tabular}{l|cc}
\toprule
                           & Source of     & Source of            \\
Mode                       &           EOS &           wordpieces \\
\midrule
1. 1st pass                & Decoder 1     & Decoder 1            \\
2. EOS2-segmented 2nd pass & Decoder 2     & Decoder 2            \\
3. EOS1-segmented 2nd pass & Decoder 1     & Decoder 2            \\
\bottomrule
\end{tabular}
}
\vspace{-0pt}
\end{table}

We again emphasize that Modes 1 and 2 are effectively single-pass models,
since all the tokens come from one decoder in both cases,
and are most representative of how ASR models are typically operated.
Mode 3 is a very uncommon approach proposed recently in \cite{huang2022e2esegmentation},
but we include it in our results for completeness.
In Section \ref{sec:results}, we study the impact of our semantic segmenter in all three modes.

\subsection{Beam search}
\label{sec:beamsearch}
We use a frame-synchronous beam search with a beam size of 4 for the first pass and 8 for the second pass, with a pruning threshold of 5 for both passes. At each frame, a breadth-first search is conducted to identify possible expansions with a limited search depth of 10. The production streaming client can only process segments up to 65 seconds. If any segment exceeds this length, it is forcibly finalized.

\subsection{Voice activity detector}
The pipeline we use incorporates a small LSTM voice activity detector \cite{zazo2016feature} at the start of the E2E model. This detector not only serves as a segmenter in our VAD segmenter experiments, but also functions as a filter for frames. Following the initial 200 ms of silence, any successive frames classified as silence are filtered out. To replicate edge deployment conditions, frame filtering is enabled for all experiments.

\section{Results}
\label{sec:results}

\begin{table*}
\caption{Segment lengths (SL), EOS latencies (EOS), and WERs on YT\_LONG. Operating modes described in Section \ref{sec:modes}.}
\vspace{-5pt}
\label{tab:main}
\centering
\resizebox{.75\textwidth}{!}{%
\begin{tabular}{l|cc|cc|ccc}
\toprule
                                                         &                    &                     &                             &                             & Mode 1                            & Mode 2                            & Mode 3             \\
Segmenter                                                & SL50               & SL90                & EOS50                       & EOS90                       &          WER                      &          WER                      &          WER       \\
\midrule                                 
E1: No segmenter                                         & \cg{65.0}{1.3}{65} & \cg{65.0}{3.0}{65}  & -                           & -                           & \cg{19.00}{18.13}{19.39}          & \cg{18.85}{16.48}{22.26}          & \cg{18.78}{15.86}{19.60}          \\
E2: Fixed-3s                                             & \cg{ 3.0}{1.3}{65} & \cg{ 3.0}{3.0}{65}  & -                           & -                           & \cg{19.20}{18.13}{19.39}          & \cg{22.26}{16.48}{22.26}          & \cg{19.60}{15.86}{19.60}          \\
E3: Fixed-5s                                             & \cg{ 5.0}{1.3}{65} & \cg{ 5.0}{3.0}{65}  & -                           & -                           & \cg{19.39}{18.13}{19.39}          & \cg{20.20}{16.48}{22.26}          & \cg{17.78}{15.86}{19.60}          \\
E4: Fixed-10s                                            & \cg{10.0}{1.3}{65} & \cg{10.0}{3.0}{65}  & -                           & -                           & \cg{19.16}{18.13}{19.39}          & \cg{18.55}{16.48}{22.26}          & \cg{16.82}{15.86}{19.60}          \\
E5: VAD                                                  & \cg{ 3.3}{1.3}{65} & \cg{14.0}{3.0}{65}  & \cg{380}{180}{380}          & \cg{490}{480}{510}          & \cg{19.13}{18.13}{19.39}          & \cg{16.64}{16.48}{22.26}          & \cg{16.23}{15.86}{19.60}          \\
E6: E2E-PauseDetector \cite{huang2022e2esegmentation}    & \cg{ 6.9}{1.3}{65} & \cg{20.8}{3.0}{65}  & \cg{240}{180}{380}          & \bg{480}{480}{510}          & \cg{18.72}{18.13}{19.39}          & \cg{16.62}{16.48}{22.26}          & \bg{15.86}{15.86}{19.60}          \\
E7: E2E-Semantic (this work)                             & \cg{ 1.3}{1.3}{65} & \cg{ 3.6}{3.0}{65}  & \bg{180}{180}{380}          & \cg{510}{480}{510}          & \bg{18.13}{18.13}{19.39}          & \bg{16.48}{16.48}{22.26}          & \cg{16.38}{15.86}{19.60}          \\
$\Rightarrow$\;\; E7 vs. E6 & - & - & \textbf{\textcolor{ForestGreen}{-60}} & \textcolor{red}{+30} & \textbf{\textcolor{ForestGreen}{-3.2\%}} & \textbf{\textcolor{ForestGreen}{-0.8\%}} & \textcolor{red}{+3.3\%} \\
\bottomrule
\end{tabular}
}
\vspace{-5pt}
\end{table*}

\subsection{LM intrinsic segmentation quality}
Before using the LM as a teacher,
we first ensure that its intrinsic segmentation quality is acceptable.
On the C4 held-out data, the LM teacher achieves a label accuracy of 97.5\% and full sequence accuracy of 74.4\%.
On the YT\_LONG test set, which has punctuation, we convert the terminal punctuation marks into \eos{} and use them as the ground-truth for measuring the teacher model's quality. The segmentation \(F_1\) score is 62.1\%.
But an error analysis showed that some of the errors are acceptable.
For example, interjections like "all right" are followed by punctuation in the reference but not separated by the model.
For comparison, a T5 base model \cite{raffel2020} fine-tuned on YouTube TED talks \cite{mccarthy2022} obtains 63.7\% \(F_1\).

\subsection{Metrics}

We now proceed to evaluate our teacher-distilled segmenter on ASR.
The metrics tracked for each experiment include
\begin{itemize}[leftmargin=*]
    \item \textbf{Segment length (SL)} in seconds, including the 50th (SL50) and 90th (SL90) percentiles.
    \item \textbf{End-of-segment (EOS) latencies} in milliseconds, including the 50th (EOS50) and 90th (EOS90) percentiles,
    which represent the time from the end of the last word to the last segment boundary.
    We report the segment lengths and latencies from when Decoder 1 is used for EOS,
    since Decoder 2 EOS is only used in latency-insensitive applications (Section \ref{sec:modes}).
    \item \textbf{WER} measures ASR quality. We measure WER for all three operating modes discussed in Section \ref{sec:modes}.
\end{itemize}

\subsection{Main results}
\label{sec:mainresults}

In Table \ref{tab:main}, we compare our model (E7) against a variety of available segmenters (E1-E6).
The base ASR model and beam search algorithm are kept fixed across all experiments.


The first baseline, no segmenter (E1), obtains a WER of around 19\% for 1st and both 2nd passes.
Adding fixed-length segmentation (E2-E4) hurts the Mode 1 and Mode 2 WERs,
which is to be expected since it may cut off in the middle of the spoken word.
Surprisingly, Mode 3 WER improves with fixed-length segmentation.
This may be an effect of the right context dummy frames (technique discussed in \cite{huang2022e2esegmentation});
by adding some dummy frames, they provide a chance for the decoder state to
correctly decode the wordpieces on both sides of a word
whenever that word is cut off.

The VAD (E5) and E2E-PauseDetector (E6) segmenters achieve much better performance than the fixed-length (E2-E4) and no-op (E1) segmenters,
providing evidence that silence-based segmentation decisions can improve over data-agnostic decisions or no segmentation at all.

Finally, the E2E-Semantic segmenter (E7) achieves further gains.
It achieves a 3.2\% relative reduction for 1st pass WER (Mode 1)
and 0.8\% relative reduction for EOS2-segmented 2nd pass WER (Mode 2)
compared to the next best result from E2E-PauseDetector (E6).
As discussed in Section \ref{sec:modes},
Modes 1 and 2 are most representative of how common ASR models are operated.
We emphasize that the \textit{only} difference between E7 and E6 is the teacher used to distill the segmenter during training,
providing controlled evidence that semantic-based EOS training signals are superior to acoustic-silence-based signals.

We further see a reduction in the median EOS latency (EOS50) from 240 ms (E6) to 180 ms (E7).
The reduction can be explained as follows:
E2E-Semantic is trained to attend mostly on text features,
while E2E-PauseDetector is trained to attend mostly to acoustic pauses;
This allows E2E-semantic to emit EOS immediately after the last word of a sentence is decoded,
whereas E2E-PauseDetector must typically wait for several more frames to ensure that the pause is long enough.
Human-computer interaction (HCI) guidelines dictate that control tasks like keystroke feedback be in the 50-150 ms range \cite{attig2017system}.
Our improved latency, which moves closer to this range,
could thus have a steep positive impact on median user experience
for, e.g., voice-control tasks in the smart assistant setting.

We also observe some limitations with the E2E-Semantic segmenter.
Most notably, we see a 3.3\% relative regression in Mode 3 WER.
A possible explanation for this requires a deeper understanding the dummy frame injection technique \cite{huang2022e2esegmentation};
Briefly, the quicker latency afforded by E2E-Semantic may cause the injected dummy frames
(which are repeats of the frame at the EOS timestamp) to be a non-silence frame,
causing the decoder to hallucinate wordpieces---this hypothesis is backed by the fact we increased error mostly from insertions.
We also see a small EOS latency degradation on the 90th percentile,
suggesting that user experience may be slightly worse in long-tail use cases.

\subsection{Ablation study}
\begin{figure}
\vspace{-5pt}
  \centering
  \includegraphics[width=.9\columnwidth]{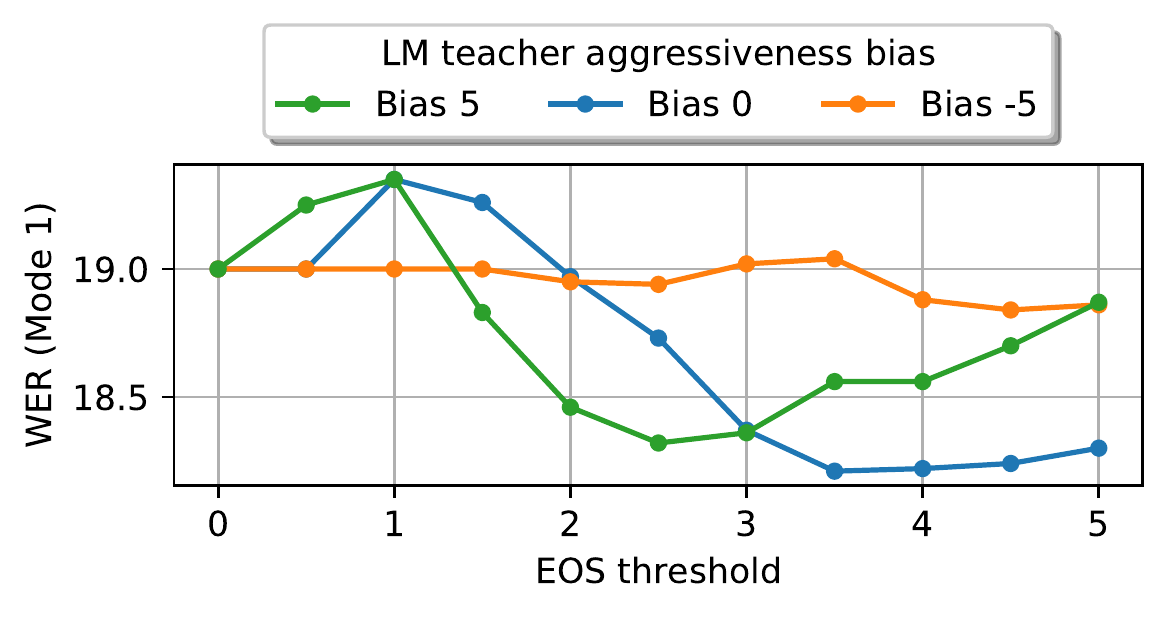}
\vspace{-9pt}
  \caption{
  Ablation study on the \eos{} threshold during distillation and inference.
  During distillation, a linear bias (colors) is added to the LM teacher's logits;
  higher bias implies more \eos{} insertions to the ground truth targets,
  making the student segmenter more aggressive.
  During inference, EOS costs below the EOS threshold (x-axis) get emitted;
  higher threshold implies more \eos{} emissions.
  }
  \label{fig:ablation}
\vspace{-15pt}
\end{figure}

We perform an ablation study where we vary both the
EOS aggressiveness of the LM teacher during training as well as the EOS threshold during inference.
We expect that higher LM teacher aggressiveness biases (more likely to emit EOS)
would need to be balanced out by tighter (lower) EOS thresholds during inference.
We observe exactly that effect in Figure \ref{fig:ablation}.
For an aggressiveness bias of 5 (green), the minimum WER occurs at an EOS threshold of 2.5.
The EOS threshold loosens to 3.5 for a less aggressive bias of 0 (blue),
and loosens further to 4.5 for an aggressiveness bias of -5 (orange).
\wrh{haozhang and shankarkumar could you write a 1-2 sentence explanation of what the aggressiveness bias is?}

\section{Conclusion}
\label{sec:conclu}
We proposed segmenting long-form speech using semantic sentence boundaries
and tested this hypothesis via distilling a bidirectional teacher LM
to a unidirectional, streaming segmenter model,
demonstrating WER and latency gains in common ASR operating modes.

\bibliographystyle{IEEEtran}
\bibliography{bibliography}

\end{document}